\def\year{2021}\relax
\documentclass[letterpaper]{article} 
\usepackage{aaai21}  
\usepackage{times}  
\usepackage{helvet} 
\usepackage{courier}  
\usepackage[hyphens]{url}  
\usepackage{graphicx} 
\usepackage{natbib}  
\usepackage{caption} 
\usepackage{amsmath}
\usepackage{booktabs}
\usepackage{graphicx}
\usepackage{soul}
\usepackage{microtype}
\usepackage{xcolor}

\title{EQG-RACE: Examination-Type Question Generation}

\author{
Xin Jia, Wenjie Zhou, Xu Sun, Yunfang Wu\thanks{\quad Corresponding author.}\\
}
\affiliations{
    MOE Key Lab of Computational Linguistics, School of EECS, Peking University \\
    \texttt{\{jemmryx, wjzhou013, xusun, wuyf\}@pku.edu.cn}\\
}

\begin{document}

\maketitle

\begin{abstract}
Question Generation (QG) is an essential component of the automatic intelligent tutoring systems, which aims to generate high-quality questions for facilitating the reading practice and assessments. However, existing QG technologies encounter several key issues concerning the biased and unnatural language sources of datasets which are mainly obtained from the Web (e.g. SQuAD).  In this paper, we propose an innovative Examination-type Question Generation approach (EQG-RACE) to generate exam-like questions based on a dataset extracted from RACE. Two main strategies are employed in EQG-RACE for dealing with discrete answer information and reasoning among long contexts. A Rough Answer and Key Sentence Tagging scheme is utilized to enhance the representations of input. An Answer-guided Graph Convolutional Network (AG-GCN) is designed to capture structure information in revealing the inter-sentences and intra-sentence relations. Experimental results show a state-of-the-art performance of EQG-RACE, which is apparently superior to the baselines. In addition, our work has established a new QG prototype with a reshaped dataset and QG method, which provides an important benchmark for related research in future work. We will make our data and code publicly available for further research. 

\end{abstract}

\section{Introduction}
\label{intro}
Automatic Question Generation (QG) aims at generating grammatical questions for a given text, which is receiving an increasing research interest in the NLP community. Besides enhancing question answering (QA) systems \cite{Tang2017QuestionAA,Duan2017QuestionGF,Xu2019AskingCQ,Zhang2019AddressingSD} and human-machine dialog generation \cite{Wang2018LearningTA}, an important purpose of QG is to generate questions of educational materials for reading practice and assessment. To facilitate the instructing process, QG has been investigated for many years \cite{Kunichika2001AutomatedQG,Mitkov2003ComputeraidedGO,Heilman2010GoodQS}.


\begin{table*}[ht]
    \centering
    \scalebox{1.0}{
    \begin{tabular}{l|l|ccc}
         \hline
         \bf Styles & \bf Examples & \bf Train & \bf Dev & \bf Test  \\
         \hline
         \hline
         Cloze & The last sentence in the passage shows that \_\_\_ . & 46075 & 2575 & 2640 \\
         General & What would be the best title for the passage ? & 23290 & 1277 & 1344 \\
         Specific & Why did Tommy's parents send him to a catholic school ? & \bf 18501 & \bf 1035 & \bf 950 \\
         \hline
         
    \end{tabular}
    }
    \caption{Different types of questions in RACE. In constructing our EQG-RACE dataset, we remove Cloze and General style questions and only maintain Specific questions. The right part shows data distribution in the original RACE dataset.}
    \label{EQG-RACE}
\end{table*}
\begin{table*}[h]
    \centering
    \scalebox{1.0}{
    \begin{tabular}{l|cccccc}
    \hline
        Dataset \quad \quad \quad \quad \quad & Passage.N & Passage.L & Question.N & Question.L & Answer.L & Vocab.N \\
    \hline
    \hline
        SQuAD & 20958 & 134.8 & 97888 & 11.31 & 2.91 & 230399  \\
        EQG-RACE & 12743 & 263.3 & 20486 & 10.87 & 6.36 & 173171 \\
    \hline
    \end{tabular}
    }
    \caption{Statistics of the most common used QG dataset SQuAD and our reconstructed EQG-RACE. N and L represent the number and average length respectively.}
    \label{lengths}
\end{table*}

Existing methods for QG mainly adopt web-extracted QA datasets, such as SQuAD \cite{Rajpurkar2016SQuAD10}, MS-MARCO \cite{Nguyen2016MSMA}, NewsQA \cite{Trischler2016NewsQAAM} and CoQA \cite{Reddy2018CoQAAC}, which are not ideal for generating questions in real context. These datasets are either domain-specific or mono-styled (such as news stories for NewsQA and Wikipedia articles for SQuAD and MS-MARCO). Answers in such datasets are often short texts extracted from the context passages and questions are automatically extracted from the Web or generated by crowd-workers. Another dataset, LearningQ \cite{Chen2018LearningQAL} is an answer-unaware QG dataset, where questions are subjective learner-generated posts extracted from E-learning videos. Different from these datasets, RACE \cite{Lai2017RACELR} is collected from English exams for Chinese students, which is a high-quality examination dataset on reading comprehension in real context. To evaluate learners' cognitive levels in reading, RACE contains articles of diversified genres (such as stories, ads, news, etc.) and questions of various levels. To generate exam-like questions in RACE, models need more advanced abilities of summarization and reasoning.   

Most QG systems usually perform sequence-to-sequence generation \cite{Du2017LearningTA,Zhou2017NeuralQG,Sun2018AnswerfocusedAP,Zhang2019AddressingSD} with attention mechanism. To generate answer-focused questions, features including answer position, POS and NER are used for encoding the contexts \cite{Zhou2017NeuralQG,Song2018LeveragingCI}. Some works also explore Copy or Pointer mechanism \cite{Hosking2019EvaluatingRF,See2017GetTT,Zhao2018ParagraphlevelNQ} to overcome the OOV problem.

In this paper, we propose Examination-type Question Generation on RACE (EQG-RACE). 
We clean the RACE dataset and maintain Specific-style questions to construct an examination-type QG dataset. According to our preliminary experiments, current QG models that perform well on SQuAD show much inferior performance on EQG-RACE, indicating the domain-transfer problem of generating examination-type questions.

As a real-world examination data designed by educational experts, there are two main factors that make EQG-RACE more challenging. First, answers are often complete sentences (or long phrases) rather than short text spans contained in the input sequences, making the previous answer tagging method invalid. Second, the context passages are longer and the questions are created through deep reasoning in multiple sentences, making the sequential encoding method like LSTM dysfunction. To address the first issue, we employ a distant-supervised method to find key answer words and key sentences, and then incorporate them into word representations. To tackle the second problem and model the reasoning relations within and across sentences, we design an Answer-guided Graph Convolutional Network(AG-GCN) to capture structure information. We conduct a series of experiments on our reconstructed EQG-RACE dataset to explore the performance of existing QG methods and the results validate the effectiveness of our proposed strategies. 



\section{Data Construction}
In the original RACE dataset, each sample consists of one passage, one question and four options (one right answer and three distractors). For constructing our EQG-RACE dataset, we process the original RACE dataset to suit the QG task.  


First, we discard distractors and only maintain the right answer for each sample. Distractors are designed to confuse learners and would introduce noises when generating a grammatical question that can be exactly answered by the right answer. 

Second, we filter some types of questions. The questions in RACE can be roughly divided into three types: Cloze, General and Specific. Cloze style questions are in the format of declarative or interrogative sentences with some missing parts to be filled by right answers, which are inappropriate for QG task. General-style questions(e.g. ``What would be the best tile for the passage?") are not specific to a particular article, but are widely applicable to any context passages, which can be generated using rule-based methods. Specific-style questions are semantically related to some specific content of the article, which are the focus of our model. Therefore, during reconstructing EQG-RACE we remove Cloze and Gneraral style questions through string matching and hand-crafted rules, and only maintain the Specific style questions. For a better understanding of our data processing, Table \ref{EQG-RACE} gives examples of three style questions as well as their statistics. 

As a result, in our EQG-RACE dataset each sample is in the form of $<$passage, answer, question$>$. Our task is to automatically generate questions given passage and answer. Two examples of EQG-RACE are given in Table \ref{case_study} for a case study.     

Furthermore, Table \ref{lengths} gives a descriptive comparison of the most commonly used QG dataset SQuAD 
and our EQG-RACE. Overall, the scale of EQG-RACE is much smaller than SQuAD (20,486 vs. 97,888 questions). The average passage length of EQG-RACE is 263.3, almost twice the length of SQuAD passage. More importantly, answers in EQG-RACE are generated by human experts and are not simply extracted from the input text. Besides, the answer length of EQG-RACE is more than twice that of SQuAD. These comparisons of textual properties show that EQG-RACE is a more challenging dataset for QG.

The original RACE contains 87,866, 4,887 and 4,934 samples for training, development and testing, respectively. After filtering, the EQG-RACE contains 18,501, 1,035 and 950 $<$passage, answer, question$>$ triples \footnote{Data and code available at: https://github.com/jemmryx/EQG-RACE}, as shown in Table \ref{EQG-RACE}. 

\begin{figure*}[t]
    \centering
    \includegraphics[width=2.0\columnwidth]{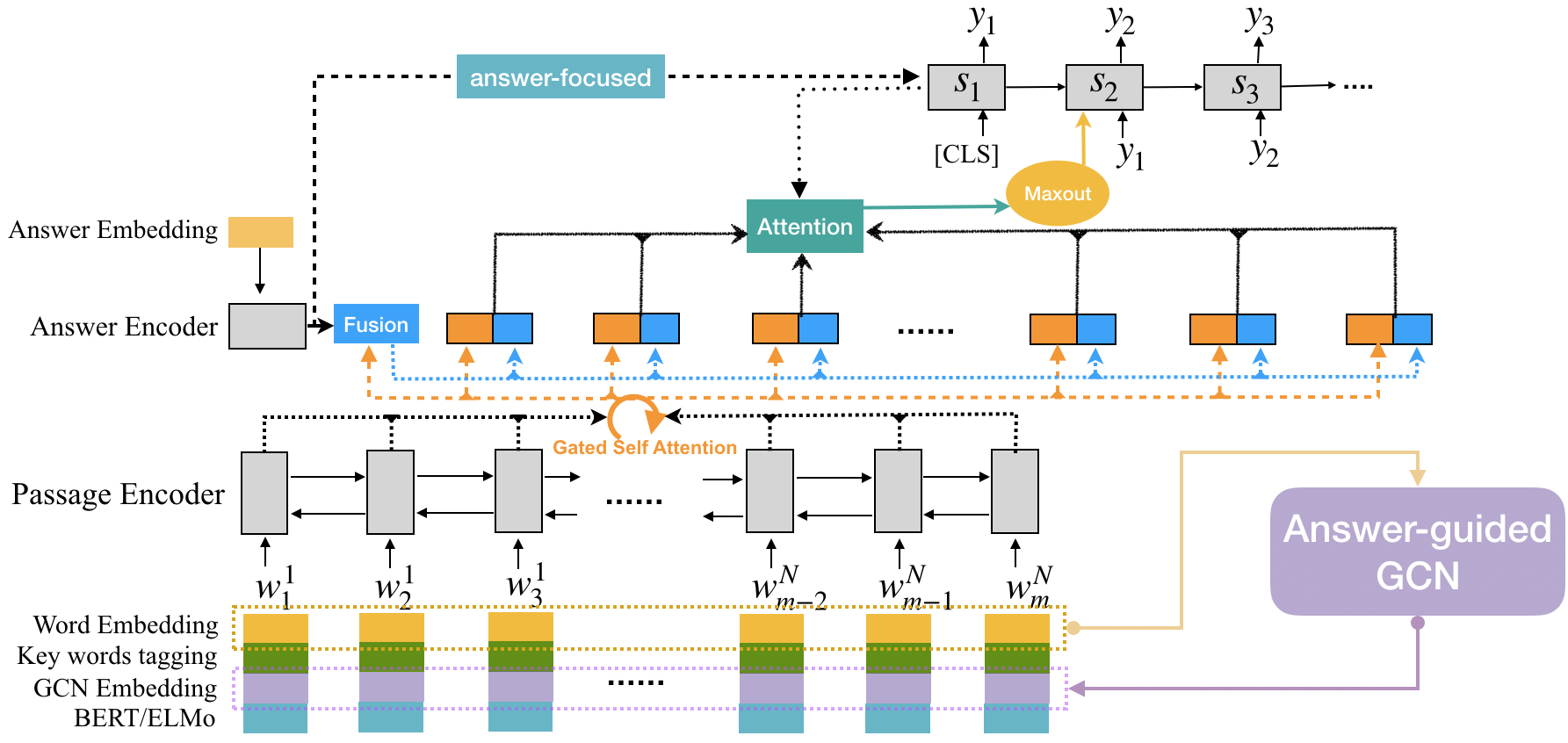}
    \caption{The illustration of our proposed unified model. BERT/ELMo represents using BERT or ELMo pre-trained embedding. (Best viewed in color)}
    \label{model}
\end{figure*}

\section{Model Description}
On the EQG-RACE data, given a passage $p$ and its corresponding answer $a$, our task aims to generate a grammatical and answer-focused question $q$:
\begin{align}
    q = \arg\max_qP(q|p,a)
\end{align}

To handle this challenging task, we propose a unified model by leveraging keywords information, as illustrated in Figure \ref{model}. First of all, we annotate passage keywords according to answer information. The input passage is fed into an answer-guided GCN to obtain answer-focused context embedding. Then, features of word embeddings, Keywords tagging embeddings, GCN embeddings and pre-training embeddings are concatenated as the input of a bidirectional LSTM encoder. A gated self-attention mechanism is then applied to the passage hidden states. Upon the above steps, we fuse passage and answer hidden states to obtain answer-aware context representations. 
Finally, an attention-based decoder generates question in sequence with the help of maxout-pointer mechanism. 

\subsection{Baseline Model}
We take the gated self-attention maxout-pointer model \cite{Zhao2018ParagraphlevelNQ} as the baseline model. 

For the encoder, we use two-layer bi-directional LSTMs to encode the input passage and get its hidden representations $H$. At each time step $t$: 
\begin{align}
    h^p_t = LSTM(h^p_{t-1}, e^p_t)
    \label{embedding_input}
\end{align}
where $h^p_t$ and $h^p_{t-1}$ are LSTM hidden states, and $e^p_t$ is word embedding. A gated self-attention mechanism \cite{Wang2017GatedSN} is then applied to $H$ to aggregate intra-passage dependencies for $\hat{H}$:
\begin{align}
    s^p_t &= H*softmax(H^TW^sh^p_t)\\
    f^p_t &= tanh(W^f[h^p_t, s^p_t])\\
    g_t &= sigmoid(W^g[h^p_t,s^p_t])\\
    \hat{h^p_t} &= g_t*f^p_t + (1-g_t)*h^p_t
    \label{encoder_hidden}
\end{align}
We obtain self-attention context vector $f^p_t$ by conducting self-matching mechanism on $H$. Then, a learnable gate $g_t$ is used to balance how much $f^p_t$ and $h^p_t$ will contribute to the encoder output $\hat{H}$.


The decoder is another two-layer uni-directional LSTM. At each time step the decoder state $d_{t+1}$ is generated according to context vector $c_t$, which aggregates $\hat{H}$ through attention mechanism:
\begin{align}
    \alpha_t  &= softmax(\hat{H}^\mathsf{T}W_ad_t)\\
    c_t &= \hat{H}\alpha_t\\
    \hat{d_{t}} &= tanh(W_c[c_{t},d_{t}])\\
    d_{t+1} &= LSTM([y_{t},\hat{d_{t}}])
\end{align}
The probability of a target word $y_t$ is computed by the maxout-pointer mechanism:
\begin{align}
    p_{vocab} &= softmax(W^ed_t)\\
    p_{copy} &= \max \limits_{where\, x_k=y_t}{\alpha_{t,k}},\quad \quad y_t\in X\\
    p(y_t|y_{<t}) &= p_{vocab}*g_p + p_{copy}*(1-g_p)
\end{align}
where $X$ is the vocab of the input sequence and $g_p$ is a trainable parameter to balance $p_{vocab}$ and $p_{copy}$. $p_{vocab}$ and $p_{copy}$ represent the probability of generating a word from vocab and copying a word from the input sequence respectively.

\subsection{Keywords Tagging}
The questions and answers in RACE are generated by human experts and the answers are not continuous spans in the context, which is different from other common used QG datasets. To this end, the traditional answer tagging methods in QG \cite{Zhou2017NeuralQG,Yuan2017MachineCB,Zhao2018ParagraphlevelNQ} can not be directly used in our task. To introduce answer-focused information into context representations, we employ Keywords tagging methods to locate answer-related words. 
\paragraph{Rough Answer Tagging} Denote answer words (words occurring in the answer text) as $A_t$. We first remove stop words in $A_t$ with NLTK to obtain meaningful content
words $A_f$. Then, we match each passage words with $A_{f}$ and label these matching words with ``A" tags. 

\paragraph{Key Sentence Tagging} Given a context passage, the related questions and answers usually focus on a specific topic which relies on one or several key sentences rather than the full passage. To capture this important information, we find the key sentence in a passage and tag all the words in this sentence with a special label ``S". Inspired by the work of \citet{Chen2018FastAS}, for each answer text $A_i$ in EQG-RACE triples, we find the most similar context sentence $S_j$ through:
\begin{align}
    j = \mathop{\arg\max}_{t}(\mathrm{ROUGE\!-\!L}_{recall}(S_t, A_i))
    \label{sentence_seletion}
\end{align}
where $S_t$ is the $t-th$ sentence in the input passage.

Note that the answer words tag ``A" has a higher priority than ``S", which means if a word has both ``A" and ``S" tags, it will be marked as ``A". If a passage word doesn't belong to any answer words nor key sentence words, it will be tagged with ``O". 

In our experiment, ``A", ``S" and ``O" tags are randomly initialized to 32-dimensional trainable variables and serve as features to enhance context representations. We concatenate Keywords tagging $k^p_t$ with word embeddings $e^p_t$ as feature-enriched input. The Equation \ref{embedding_input} can be rewritten as:
\begin{align}
    h^p_t = LSTM(h^p_{t-1}, [e^p_t; k^p_t])
    \label{tagging_input}
\end{align}

\subsection{Answer-guided Graph}
The questions in RACE are designed to test a learner's understanding ability (such as summarization and attitude analysis). Generating such high-quality questions involves varying cognitive skills and often requires deep reasoning of complex relationships both intra-sentence and inter-sentences. This is different from the existing QG dataset like SQuAD that mainly accommodates factual details in context. As a result, the traditional CNN and RNN methods cannot meet the demands of this difficult task.

\begin{figure}[t]
    \centering
    \includegraphics[width=0.95\columnwidth]{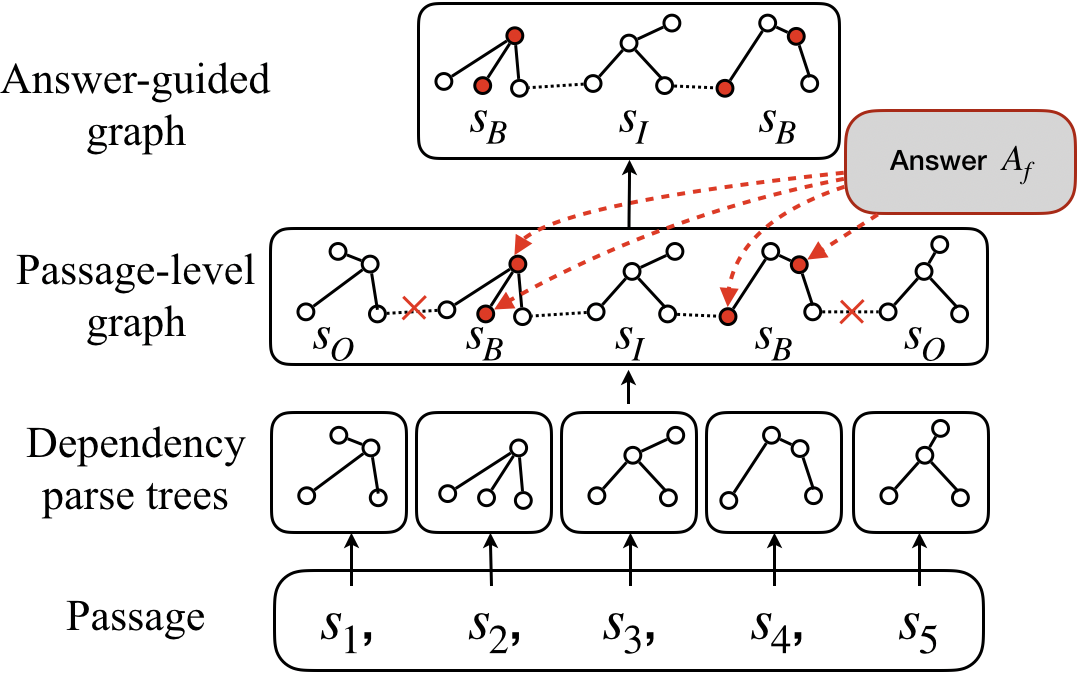}
    \caption{Construction of our Answer-guided graph. After obtaining key answer words $A_f$, the passage is denoted as: [$S_O$, $S_B$, $S_I$, $S_B$, $S_O$]. We delete two $S_O$ sentences since they do not contain answer words and not connect any two $S_B$.}
    \label{graph}
\end{figure}

To address this issue, we propose an Answer-guided Graph Convolution Network (AG-GCN) to encode passages. As illustrated in Figure \ref{graph}, the graph construction can be elaborated through the following steps:
\begin{itemize}
    \item \textbf{Step 1}: Implementing dependency parsing for each sentence in the context passage.
    \item \textbf{Step 2}: Linking neighboring sentences' dependency trees by connecting nodes which are at sentence boundaries and next to each other, to build a passage-level dependency parse graph.
    \item \textbf{Step 3}: Retrieving the key answer words $A_f$ for a given answer through Rough Answer Tagging method and removing the ``isolated" nodes and their edges from the passage-level graph to construct Answer-guided graph.
\end{itemize}

After Step 2, we obtain a passage-level dependency parse graph and each passage word corresponds to a node in the graph. However, not every word entitles to generate an answer-focused question, but only the key answer words and its related terms contribute to the QG process. To reduce redundant information and focus on key words, we remove the ``isolated" nodes and their edges from the passage graph. Specifically, we divide sentences in the passage into 3 groups according to key answer words $A_f$: $S_B(Begining)$ represents sentences containing words in $A_f$; $S_I(Inside)$ represents sentences that connect any two important sentences $S_B$; others are represented as $S_O(Out)$. We consider the nodes in $S_O$ as ``isolated" nodes since they do not contain answer-focused information and can not contribute to the reasoning process between two essential sentences.


The unweighted graph adjacency matrix can be denoted as $A$, where the weights $A_{ij}$ is 1 if node $i$ is connected with node $j$ otherwise is 0. For the implementation of GCN, we follow the work of \citet{Kipf2016SemiSupervisedCW}. We take $A$ and word embeddings $E$ as input, and the encoding process can be formulated as:


\begin{align}
    g_t^{l+1} &= \sigma(\Tilde{D}^{-\frac{1}{2}}\Tilde{A}\Tilde{D}^{-\frac{1}{2}}g_t^lW^l)\\
    \Tilde{D}_{ii} &= \sum_j\Tilde{A}_{ij}\\
    \Tilde{A} &= A+I_N
\end{align}
where $D$ is a diagonal matrix and $I_N$ is the identity matrix. The $l$-th layer of GCN takes the last layer's output $g_t^{l-1}$ as input and $g_t^0$ is set to $E$. We adopt a two-layer GCN model.

The output of the last layer $g_t^L$ is fed into a feed-forward layer to obtain the final output $g^p_t$. We then concatenate it with the word embedding and Keywords tagging embedding as the encoder inputs. Accordingly, Equation \ref{tagging_input} can be rewritten as:

\begin{align}
        h^p_t = LSTM(h^p_{t-1}, [e^p_t;k^p_t;g^p_t])
\end{align}
\begin{table*}[t]
    \centering
    \scalebox{1.0}{
    \begin{tabular}{l|cccccc|c}
    \hline
    & \multicolumn{6}{c|}{EQG-RACE} & \multicolumn{1}{c}{SQuAD} \\
    \hline
        \bf Previous Models \quad \quad \quad \quad \quad \quad & BLEU-1 & BLEU-2 & BLEU-3 & BLEU-4 & ROUGE-L & METEOR & BLEU-4 \\
    \hline
    \hline
         Seq2Seq  & 23.46 & 10.24 & 6.39 & 4.75 & 23.82 & 8.57& 15.16  \\
         Pointer-generator  & 28.96 & 14.33 & 8.82 & 5.99 & 30.02 & 12.26 & 15.21 \\
         HRED  & 28.69 &15.14 &9.42 & 6.16 & 32.70 & 12.48 &16.43 \\
         Transformer & 28.93 & 15.20 & 9.45 & 6.25 & 32.43 & 13.49 & 16.50 \\
         ELMo-QG  & 33.95 & 18.55 & 11.93 & 8.23 & 33.26 &14.35&16.75\\
    \hline
    \hline
    \bf Our Model &\multicolumn{6}{c|}{} & \multicolumn{1}{c}{}  \\ 
    \hline
    Unified model + ELMo& \bf 35.10& \bf 21.08& \bf 15.19& \bf 11.96& \bf 34.24& \bf 14.94 & -- \\
    \hline
    \end{tabular}
    }
    \caption{Experimental results of our proposed model comparing with previous methods. We use the source codes available on the web to conduct experiments on our EQG-RACE dataset and SQuAD. Additionally, we apply our \textbf{Rough Answer Tagging} strategy to these previous models because the existing answer tagging methods are inappropriate for our EQG-RACE.}
    \label{pre_results}
\end{table*}

\subsection{Exploring Pre-training Embeddings}
Since the number of EQG-RACE samples is relatively small, the deep neural networks may encounter under-training situations. To solve this problem 
we take pre-training embeddings $p^p_t$, such as BERT \cite{Devlin2019BERTPO} and ELMo \cite{Peters2018DeepCW} embedding, as supplements to the encoder input:
\begin{align}
        h^p_t = LSTM(h^p_{t-1}, [e^p_t;k^p_t;g^p_t;p^p_t])
\end{align}

Here, we obtain the passage encoding $h^p_t$ through a BiLSTM network. Further, we conduct self-attention operation on $h^p_t$ to get high-level representations $\hat{h^p_t}$, by Equation \ref{encoder_hidden}.   

\subsection{Passage-answer Fusion}
To well capture the inter-dependencies between passage $P$ and answer $A$,
we fuse the answer representation $h^A$ with passage representation $\hat{h^p_t}$ to achieve answer-aware representations as the final encoder outputs:  

 
\begin{align}
    \Tilde{h^p_t} = tanh(W^u[\hat{h^p_t}; {h^A}; \hat{h^p_t}*{h^A};\hat{h^p_t}+{h^A}])
\end{align}
where $h^A$ is the answer hidden state obtained via a bi-directional LSTM network.

Besides, in the decoding procedure, the first interrogative word is one of the most essential part of the whole generated question. 
Therefore, instead of using the last hidden state $\hat{h^p_t}$ of the passage encoder, we utilize the answer encoder states $h^A$  as the initialization of the decoder \cite{Kim2018ImprovingNQ}. Under this setting, the decoder is likely to generate more answer-focused interrogative words.


\section{Experiments and Results}

\begin{table*}[t]
    \centering
    \scalebox{1.0}{
    \begin{tabular}{l|cccccc}
    \hline
        \bf Methods & BLEU-1 & BLEU-2 & BLEU-3 & BLEU-4 & ROUGE-L & METEOR  \\
    \hline
    \hline
    Maxout-pointer(no answer information) & 27.21 & 12.51 & 6.97 & 4.06 & 26.57 & 10.72  \\
    \hline
    Unified model & \bf 36.22& 20.41& 13.26& 9.09& \bf 35.74& 15.18\\
    - answer-guided-graph& 34.15& 18.44& 11.54& 7.43& 33.66& 14.25 \\
    - passage-answer-fusion& 34.37& 18.89& 11.83& 7.56& 34.62& 14.51 \\
    - key-sentence-tagging& 33.89& 18.38& 11.54& 7.62&	34.07& 14.28 \\
    - rough-answer-tagging& 32.74& 17.07& 10.27& 6.32&	32.51& 13.43\\
    \hline
    Unified+BERT& 34.92& 20.73& 14.09& 10.01&  34.63& \bf 15.54\\
    Unified+ELMo&  35.10& \bf 21.08& \bf 15.19& \bf 11.96& 34.24& 14.94 \\
    
    \hline
    \end{tabular}
    }
    \caption{Ablation study of our proposed model on EQG-RACE dataset.}
    \label{exp_results}
\end{table*}

\subsection{Experimental Settings}
In our model, the LSTM hidden sizes of encoder and decoder, the word embedding size and the GCN hidden size are all 300. We set the vocabulary to the most frequent 45,000 words. The maximum lengths of input passage and output question are 400 and 30, respectively. We use pre-trained GloVe embedding as initialization of word embedding and fine-tune it during training. We employ Adam as optimizer with a learning rate 0.001 during training.

For \textbf{Unified model + ELMo} settings, we use the pre-trained character-level word embedding from ELMo \cite{Peters2018DeepCW} as additional features and also fine-tune it during training. For \textbf{Unifed model + BERT}, we replace the ELMo with BERT and keep other settings the same. We use the WordPiece tokenizer to process each word to fit BERT and conduct post-processing to map outputs into normal words. 

We utilize Stanford CoreNLP to get dependency parse trees of sentences. The dropout rate of both encoder and GCN is set to 0.3. In decoding, the beam search size is 10.

We evaluate the performance of our models using \textbf{BLEU}, \textbf{ROUGE-L} and \textbf{METEOR}, which are widely used in previous works for QG.

\subsection{Baseline Models}
To compare our model with previous approaches, we re-implement several current neural network-based methods of text generation that have released codes on the web:

\begin{itemize}

\item \textbf{Seq2seq} \cite{Hosking2019EvaluatingRF}: A RNN sequence-to-sequence model with attention and copy mechanism.

\item \textbf{Pointer-generator} \cite{See2017GetTT}: A LSTM based seq2seq model with pointer mechanism.

\item \textbf{HRED} \cite{Gao2018GeneratingDF}: A seq2seq model with a hierarchical encoder to model the context and to capture both sentence-level and word-level information.

\item \textbf{Transformer} \cite{Vaswani2017AttentionIA}: A standard transformer-based seq2seq model.

\item \textbf{ELMo-QG} \cite{Zhang2019AddressingSD}: A maxout-pointer model with feature-enriched input which contains answer position, POS, NER and ELMo features.
\end{itemize}
The most important difference between EQG-RACE and SQuAD is that answers in SQuAD are continuous text spans while answers in EQG-RACE are not. When re-implementing these models, we utilize our proposed Rough Answer Tagging strategy to replace the previous answer tagging method and keep other settings unchanged.

\subsection{Main Results}
The experimental results are shown in Table \ref{pre_results}. The traditional text generation models that perform well on SQuAD including Seq2seq, Pointer-generator, and HRED don't work well on EQG-RACE, demonstrating the difficulty and challenging of ELMo-QG. Although the Transformer-based seq2seq model has the ability to capture global information, it does not perform well on this difficult task. Compared with previous approaches, our unified model obtains great improvements over all evaluation metrics. In particular, our method improves a 3.73 BLEU-4 over the best performance ELMo-QG that also utilizes ELMo pre-training embeddings. We achieve a state-of-the-art 11.96 BLEU-4 score on EQG-RACE, which establishes a new benchmark for future research. 

\subsection{Model Analysis}
Detailed analysis of the proposed model is presented in Table \ref{exp_results}. The maxout-pointer baseline model gets a 4.06 BLEU-4 score without using any answer information. When applying our proposed components, the unified model obtains a much better performance (especially 9.09 BLEU-4 score) on this tough EQG-RACE dataset, which is even better than ELMo-based previous model (8.23 BLEU-4 score). Based on this, we further apply pre-trained embeddings (BERT and ELMo) and achieve new state-of-the-art results on EQG-RACE.


As shown in the middle part of Table \ref{exp_results}, removing each of our proposed modules has varying degrees of decline.
Among them, dismantling Rough answer Tagging results in the biggest performance drop (-2.77), because this module highlights the positions directly related to answer which are the most essential information for generating answer-focused questions. The second most influential module is the Answer-guided Graph and removing it will cause the BLEU-4 score to decrease by 1.66. This module provides structure information both within a sentence and between adjacent sentences, which is difficult for LSTM or RNN to capture. Additionally, when removing Fusion or Key Sentence Tagging, the performance shows similar decreases.


Additionally, the pre-trained embeddings improve our model's performance with a large margin. Combining BERT embedding, we obtain a 10.01 BLEU-4 score. We achieve a state-of-the-art result of 11.96 with ELMo embeddings added to our unified model.

\subsection{Human Evaluation}
To assess the quality of generated questions, we conduct human evaluation to compare the baseline maxout-pointer model and our unified model. We randomly select 100 samples and ask three annotators to score them in the scale of [0,2] independently, with the following three metrics:
\begin{itemize}

\item \textbf{Fluency:} whether a question is grammatical and fluent.

\item \textbf{Relevancy:} whether the question is semantic relevant to the passage.

\item \textbf{Answerability:} whether the question can be answered by the right answer. 
\end{itemize}
We then compute the average value of three persons as the final score. As shown in Table \ref{human_eval}, our unified model yields higher scores on all three metrics with high Spearman coefficients. Especially, our model obtains an obvious performance gain in \textbf{Answerability}, since we incorporate answer information into the neural network in multiple ways. 

\begin{table}[h]
    \centering
    \scalebox{1.0}{
    \begin{tabular}{l|c|c|c}
    \hline
         Models & Fluency & Relevancy & Answerability  \\
         \hline
         Baseline & 1.62 & 1.45 & 0.25 \\
         Unified & \bf 1.66 & \bf 1.55 & \bf 0.71 \\
         \hline
         Spearman & 0.75 & 0.57 & 0.56 \\
         \hline
    \end{tabular}
    }
    \caption{Human evaluation results of generated questions. The baseline is the maxout-pointer model.}
    \label{human_eval}
\end{table}

\begin{table*}[t]
    \centering
    \scalebox{0.98}{
    \begin{tabular}{p{1.0\columnwidth}|p{1.0\columnwidth}}
        \hline
        \bf Passage-1: & \bf Passage-2: \\
        we have twenty minutes' break time after the second class in the morning. look! most of us are playing during the break time. \underline{some {\color{blue}{students}} are on the {\color{red}{playground}}}. they are playing basketball. oh! a boy is running with the ball. and another is stopping him. they look so cool. and there are some girls watching the game. some {\color{blue}{students}} are in the classroom. they are {\color{blue}{talking}}. a few of them are reading and doing homework. look! a girl is looking at the birds in the tree in front of the classroom. she must be thinking of something interesting because she is smiling. what are the teachers doing? some of them are {\color{blue}{working}} in the office. \underline{and some are {\color{blue}{talking}} with {\color{blue}{students}}}. everyone is doing his or her things, busy but happy! & in some countries, people eat rice every day. sometimes they eat it twice or three times a day for breakfast, lunch and supper. some people do not eat some kinds of meat. muslims, for example, do not eat pork. japanese eat lots of {\color{red}{fish}}. they are near the sea. \underline{so it is {\color{red}{easy}} for them to {\color{red}{get}} {\color{red}{fish}}}. \underline{in the west, such as england and the usa, the most important} \underline{food is {\color{blue}{potatoes}}}. people there can cook potatoes in many different ways. some people eat only fruit and vegetables. they do not eat meat or {\color{red}{fish}} or anything else from animals. they eat food only from plants. they say the food from plants is better for us than meat. \\
        \\
        \hline
        \hline
        \textbf{Answer-1}: on the {\color{red}{playground}}. & \textbf{Answer-1}: because it is {\color{red}{easy}} for them to {\color{red}{get fish}}.\\
        \textbf{Reference-1}: where are the students playing basketball? & \textbf{Reference-1}: why do japanese eat lots of fish?\\
        \textbf{Output-1}: where are the students playing basketball? & \textbf{Output-1}: why do japanese eat lots of fish?\\
        \\
        \hline
        \textbf{Answer-2}: {\color{blue}{working}} or {\color{blue}{talking}} with {\color{blue}{students}}. & \textbf{Answer-2}: {\color{blue}{potatoes}}.\\
        \textbf{Reference-2}: what are the teachers doing? & \textbf{Reference-2}: what is the most important food in some western countries?\\
        \textbf{Output-2}: what are the students working in the classroom? &  \textbf{Output-2}: what is the most important food in the usa?\\
        \hline
        \hline
    \end{tabular}}
    \caption{Case study of our generated questions. We highlight the answer words obtained by our Rough Answer Tagging in the passage (red color for answer-1 and blue color for answer-2) and underline the key sentences obtained by our Key Sentence Tagging strategy. }
    \label{case_study}
\end{table*}

\subsection{Case Study}
To illustrate our EQG-RACE task and present the output clearly, two real examples of generated questions are shown in Table \ref{case_study}. For the same passage-1, answer-1 and answer-2 focus on different aspects. The output question-1 is satisfactory because the answer word ``\textit{playground}" appears once in the passage, which allows the attention model to focus on its surrounding words to generate a right question. For answer-2, the answer words ``\textit{working}", ``\textit{talking}" and ``\textit{students}" are scattered at several positions which confuse the model's attention. 

In passage-2, the keywords of answer-1 ( ``\textit{easy}", ``\textit{get}", ``\textit{fish}") appear in several neighbouring sentences. Our Answer-guided graph may capture their inner relationship, enabling our model to produce the right question.  
As for answer-2, the answer word ``\textit{potatoes}" only appears once in the context passage. Our model also generates the right answer but is not as good as the reference question, since the expert leverages external knowledge ``\textit{usa is one of the western countries}" that does not appear in the context passage. How to employ knowledge (or common sense) for QG remains an open issue for future research.   


\section{Related Work}
Generating educational questions for reading practice and assessment is one of the most important applications of QG, which has been proposed and investigated for many years \cite{Rus2007ExperimentsOG,Rus2009The2W,Rus2011QuestionGS,Wang2018QGnetAD,Willis2019KeyPE}.
Traditional rule-based methods of QG heavily depend on the quality of handcrafted rules which are time-consuming and laborious. \cite{Wang2007AutomaticQG,Heilman2010GoodQS,Adamson2013AutomaticallyGD}. 

Recently, the neural network-based methods for automatically generating questions have achieved great success. Most of these methods treat context passages as inputs and regard questions as targets to perform sequence-to-sequence generation \cite{Du2017LearningTA,Zhao2018ParagraphlevelNQ,Zhou2019MultiTaskLW,Li2019ImprovingQG,Dong2019UnifiedLM}. To incorporate more token features, answer position and other lexical tags such as POS and NER are assimilated into the encoder \cite{Zhou2017NeuralQG, Sun2018AnswerfocusedAP, Song2018LeveragingCI,Kim2018ImprovingNQ,Chen2019NaturalQG}. To cope with long input contexts, hierarchical architecture has been used to model the contexts and capture sentence-level and word-level importance factors \cite{Gao2018GeneratingDF}. \citet{Song2018LeveragingCI} propose a matching strategy that firstly select key-sentences from long passages and perform seq2seq generation based on key-sentences. \citet{Zhao2018ParagraphlevelNQ} use a self-attention mechanism for the encoder to locate the salient information in a passage. Based on \citet{Zhao2018ParagraphlevelNQ}'s work, \citet{Zhang2019AddressingSD} apply reinforcement learning to improve the model performance; \citet{Nema2019LetsAA} utilize an answer encoder to independently encode the answer and fusion passage and answer representations. These methods don't model the structure information which is essential for reasoning among context passages in QG process.

To capture structural information, the Graph Convolutional Networks (GCN) has been employed for both text classification tasks \cite{Peng2018LargeScaleHT,Yao2018GraphCN,Liu2018MatchingLT}, and generation tasks. \citet{Zhao2018GraphSeq2SeqGF} utilize GCN to model the dependency between document words to upgrade machine translation. \citet{Li2019CoherentCG} use a topic interaction graph GCN to generate coherent comments. \citet{Chen2019NaturalQG} design both static dependency graph and dynamic attention graph GCN to perform question generation. Different from the previous work, we design an Answer-guided GCN to reduce redundant information and focus on answer-related contents.

\section{Conclusion}
We present EQG-RACE to automatically generate examination-type questions. We reconstruct the original RACE dataset to suit for question generation and will release this new data. To deal with the discrete answer information in context passages, we propose a Rough Answer and Key Sentence Tagging scheme to locate answer-related contents. Furthermore, we design an Answer-guided graph to capture answer-focused structure information as supplements for seq2seq models. Our integrated model achieves promising results on EQG-RACE and provides a solid benchmark for further research. There is still much room to improvement for this challenging task. We will explore more advanced graph structures to encode the context passages, and employ pre-trained language modeling-based methods on this task.

\section*{Acknowledgments}
This work is supported by the National Natural Science Foundation of China (62076008,61773026) and the Key Project of Natural Science Foundation of China (61936012).

\bibliography{jiaxin}
\end{document}